\documentclass[10pt,twocolumn,letterpaper]{article}

\usepackage[pagenumbers]{wacv} 

\usepackage{graphicx}
\usepackage{amsmath}
\usepackage{amssymb}
\usepackage{booktabs}

\usepackage{multirow}
\usepackage{multicol}

\usepackage[pagebackref,breaklinks,colorlinks]{hyperref}

\usepackage[capitalize]{cleveref}
\crefname{section}{Sec.}{Secs.}
\Crefname{section}{Section}{Sections}
\Crefname{table}{Table}{Tables}
\crefname{table}{Tab.}{Tabs.}

\def\wacvPaperID{767} 
\def\confName{WACV}
\def\confYear{2024}

\usepackage{xcolor}
\definecolor{airforceblue}{rgb}{0.36, 0.54, 0.66}

\newcommand{\rgbd}{\mbox{\sc{{RGB$\rightarrow$Depth}}\xspace}}

\newcommand{\zzhu}[1]{#1}

\begin{document}

\title{Consistent Multimodal Generation via A Unified GAN Framework}


\author{
  Zhen Zhu \\
  UIUC \\
  {\tt\small zhenzhu4@illinois.edu} \\
  \and
  Yijun Li \\
  Adobe Inc. \\
  {\tt\small yijli@adobe.com}
  \and
  Weijie Lyu\\
  UIUC \\
  {\tt\small wlyu3@illinois.edu} \\
  \and
  Krishna Kumar Singh \\
  Adobe Inc. \\
  {\tt\small krishsin@adobe.com}
  \and
  Zhixin Shu \\
  Adobe Inc. \\
  {\tt\small zshu@adobe.com}
  \and
  Soeren Pirk \\
  Adobe Inc. \\
  {\tt\small soeren.pirk@gmail.com}
  \and
  Derek Hoiem \\ 
  UIUC \\
  {\tt\small dhoiem@illinois.edu} \\
}

\twocolumn[{%
\renewcommand\twocolumn[1][]{#1}%
\maketitle
\begin{center}
    \centering
    \captionsetup{type=figure}
    \includegraphics[width=\textwidth]{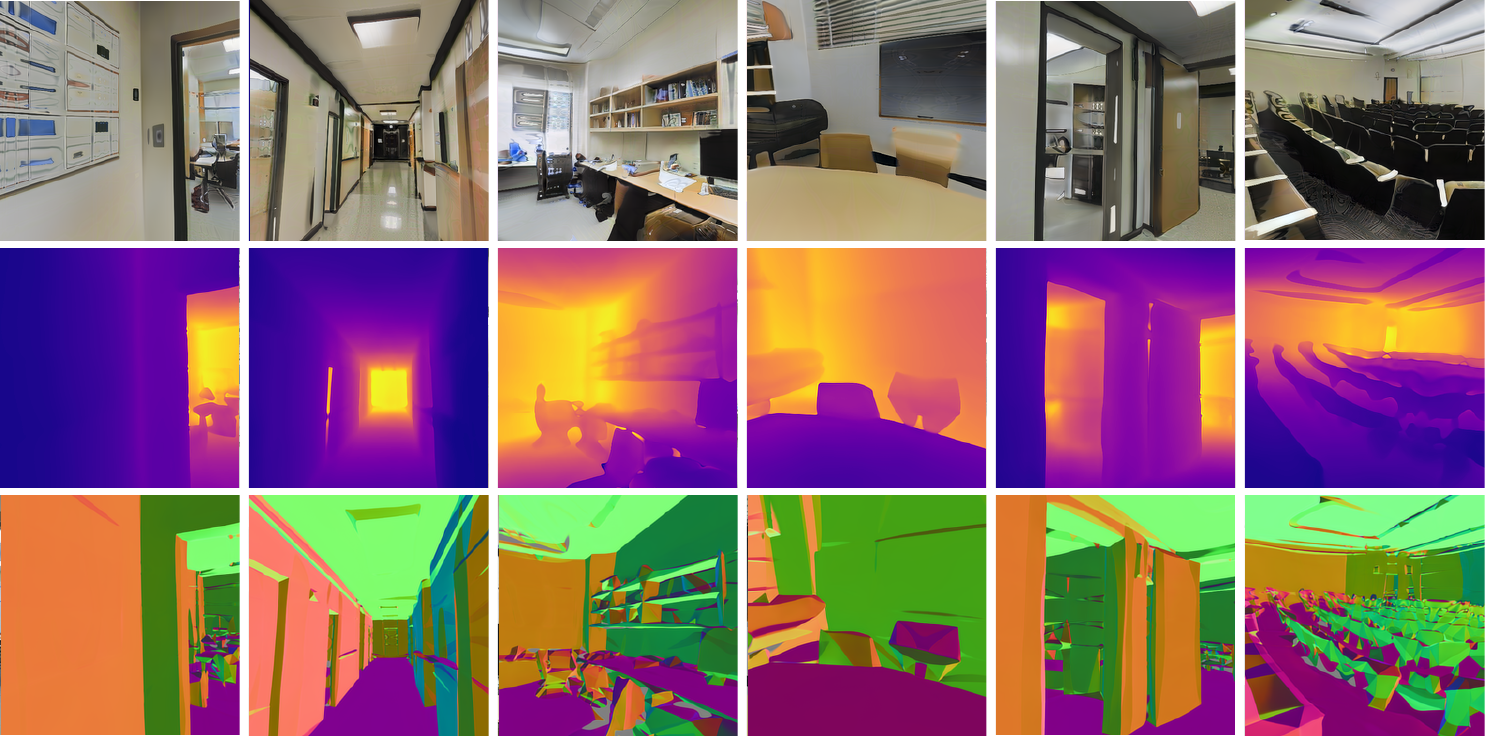}
    \captionof{figure}{Generated samples of our MultimodalGAN. Results from top to bottom row are RGB, depth and surface normals respectively. Each column is generated from the same latent code in one run and shows consistency between each modality. These samples cover diverse classes in Stanford2D3D dataset~\cite{armeni2017stanford2d3d} and exhibit multiple camera poses even without using camera parameters. \label{fig:teaser}}
\end{center}%
}]
\maketitle


\begin{abstract}
 
We investigate how to generate multimodal image outputs, such as RGB, depth, and surface normals, with a single generative model. The challenge is to produce outputs that are realistic, and also consistent with each other. Our solution builds on the StyleGAN3 architecture, with a shared backbone and modality-specific branches in the last layers of the synthesis network, and we propose per-modality fidelity discriminators and a  cross-modality consistency discriminator. In experiments on the Stanford2D3D dataset, we demonstrate realistic and consistent generation of RGB, depth, and normal images.  We also show a training recipe to easily extend our pretrained model on a new domain, even with a few pairwise data. We further evaluate the use of synthetically generated RGB and depth pairs for training or fine-tuning depth estimators. Code will be available~\hyperlink{https://github.com/jessemelpolio/MultimodalGAN}{here}.

%


\end{abstract}


\section{Introduction}
\label{sec:intro}

Realistic synthetic data is useful for training, especially when supervision is difficult to collect, such as for scene geometry. 
Ability to generate RGB images has progressed tremendously~\cite{Ian2014GAN,karras2019styleganv1,karras2020styleganv2,karras2021styleganv3}, but, without additional scene information, synthetic RGB images are of limited use for training. Our goal is to produce images in multiple modalities that are realistic and mutually consistent. Our key idea is that, just as various scene properties (e.g. RGB values and depth) stem from an underlying scene and viewpoint, we can generate them with a single model based on a shared underlying representation. 
By sharing representations among related tasks, we enable our model to generalize better on the target task. 

We build on the StyleGAN3~\cite{karras2021styleganv3} architecture. A single latent code is used to generate a shared backbone representation, which is refined by branches to generate each modality. 
%
%
Different from previous works that use certain modality-related losses, we employ only adversarial losses, which offers flexibility to scale up to many modalities without considering modality specifics. This approach eliminates the need for laborious loss weight tuning associated with modality-related losses. We introduce two types of discriminators, the fidelity and consistency discriminators, to encourage the realism and consistency among the output modalities. Furthermore, instead of specifying different data augmentation policies for different modalities, we adopt a single data augmentation strategy across all modalities, ensuring our framework's scalability and simplicity.

%

\zzhu{We demonstrate the use of our \textbf{MultimodalGAN} by applying it to a set of challenging tasks. First, we show that multiple modalities can realistically be generated together and with a high degree of consistency. 
Second, we apply our model in a multi-modality adaptation setting, where we generate modalities for a new domain. Considering that a source domain contains abundant pairs of RGB and depth images, while the target domain has abundant RGB images but very few or no depth data, our model can be used for generating non-RGB data in the target domain. 
Third, we use our framework to synthesize depth estimation dataset that comprises of RGB images and the corresponding depth maps, which can be used to train depth estimation models.
Evaluation results on the challenging Stanford2D3D~\cite{armeni2017stanford2d3d} real dataset show that our framework is capable of synthesizing more realistic multi-modality outputs compared to state-of-the-art approaches.}

%
%



\zzhu{To summarize, our main contributions include: 1) We devise an unconditional GAN framework that generates multiple modalities simultaneously. We showcase that this {MultimodalGAN} supports smooth and consistent transitions by traversing the latent space. 2) We propose a training recipe that extends the MultimodalGAN pretrained on a domain to another domain even with only a few training pairs. 3) We demonstrate our model can serve as a powerful data engine to synthesize RGB images and corresponding depth labels to improve the performance on depth estimation.}


\section{Related Work}
\label{sec:related}


\vspace{0.5em}
\noindent\textbf{Generative image modeling.}~Generative modeling by learning the data distribution has initiated a considerable amount of interests that has led to remarkable progress in recent years. Seminal approaches range from variational autoencoder~\cite{kingma2013auto} and generative adversarial networks~\cite{Ian2014GAN,brock2018large} to flow-based models and denoising diffusion probabilistic models~\cite{dhariwal2021diffusion}. Due to larger datasets and increasingly more powerful compute-availability, the quality of the results these methods generate steadily increases. However, most of these generative models are learning the distribution from the RGB modality, only to synthesize realistic images~--~other data modalities are rarely considered. 

Here we highlight a few approaches that also address other modalities in addition to generating RGB images. The work of $S^{2}$-GAN~\cite{wang2016ssgan} proposes a two-stage sequential framework by first generating surface normals and then the RGB result for indoor scenes. In contrast,  DatasetGAN~\cite{zhang2021datasetgan} focuses learning semantic segmentation maps along with RGB images in a few-shot generation scenario. After learning the GAN model on the RGB modality, it is required to label a few segmentation maps for the generated images to then fine-tune an additional network to produce segmentation labels. \zzhu{SemanticGAN~\cite{semanticGAN} also tackles this task by generating RGB and segmentation masks concurrently. It relies on an additional segmentation model to provide supervision for mask generation whereas in our case we utilize the generic adversarial loss that is agnostic to domains and do not rely on pre-trained models.}
Noguchi~\etal~\cite{noguchi2019rgbd} introduce RGBD-GAN, which is based on unsupervised 3D representation learning from 2D images, without using depth data but by relying on a 3D consistency loss. 
\zzhu{Polymorphic-GAN~\cite{PolymorphicGAN} learns morph maps on shared features, enabling style transfers only in the RGB domain. However, we focus on generating dense image analysis such as depth and surface normal.}
To make the generation more 3D-aware, a recent work~\cite{shi2022cuhk} employs two generators to synthesize RGB and depth. However, instead of using real depth data, they leverage a pretrained depth estimator~\cite{yin2021depthestimation} to extract synthetic depth as the training data, which is consequently limited by the depth range learned in the estimator. 
\zzhu{Unlike the existing approaches, we do not focus on a particular modality, but address the simultaneous generation of multiple modalities by only using one single network.}

\vspace{0.5em}
\noindent\textbf{Multi-task learning.}~It has been recognized that the joint training on multiple data modalities enables learning more powerful representations ~\cite{9156702,lu2021taskology,7989023,8100183,casser2019struct2depth}. Classic multi-task learning~\cite{crawshaw2020surveymtl,zhang2021surveymtl} aims at simulating the learning process of human beings based on the human intelligence of integrating knowledge across domains.
The general idea of multi-task learning is to train machine learning models with data from a set of related tasks simultaneously, in order to discover a shared representations. Oftentimes, configuring and training a multi-task architecture is challenging as hyperparameters and losses need to be carefully tuned jointly. The seminal approach introduced by Zamir \etal~\cite{zamir2018taskonomy} proposes a fully computational approach for modeling the structure of the space of visual tasks to quantify the relationship between them.
%
One recent work~\cite{bao2022generative} proposes to use generative approaches for visual multi-task learning, by coupling a discriminative multi-task network with a generative network. However, it aims to predict multi-task outputs from RGB images and leverages the generative network to synthesize diverse images to facilitate training.
%
\zzhu{Unlike previous works focusing on discriminative tasks, our model generates the RGB image along with its pixel-level dense representation (e.g., depth) directly from randomly sampled latent codes. 
}


\section{Method}
\label{sec:method}

\subsection{Overview}

As shown in Fig.~\ref{fig:generator}, starting from a random latent $z\in \mathcal{R}^l$ of length $l$ drawn from standard Gaussian distribution $\mathcal{N}(0, 1)$, the generator $G$ aims to convert $z$ to generate different modalities: 
\begin{equation}
    [g_m] = G(z), \ \ m\in\{\mathrm{RGB, depth, normal} \},
\end{equation}
where $[\cdot]$ indicates concatenation of outputs in the channel dimension, $g_m$ is the generation of a particular modality represented by $m$. This derives the unconditional multi-modality generator. In order to produce realistic outputs for multiple modalities, the generator should have adequate capacity. One of the most powerful unconditional generators in the literature is StyleGANs~\cite{karras2019styleganv1,karras2020styleganv2,karras2021styleganv3} which shows smooth latent interpolation and high generation quality. The structure of our generator is innovated by StyleGAN3~\cite{karras2021styleganv3}. 

A StyleGAN generator is comprised of a mapping network and a synthesis network. In StyleGAN3, the depth of the mapping network is reduced from 8 to 2. We follow the same design as it introduces less parameters and reduces the effort of optimization. The mapping network transforms the input latent $z$ to an intermediate latent space $w \in \mathcal{W}$. $w$ is passed to the affine transformation layers $A$ of all synthesis blocks in the synthesis network, adjusting the styles accomplished inside each block. 
We maintain most of the designs for the synthesis network in StyleGAN3, including Fourier features to replace the constant input, discarding noise input to ensure transformation hierarchy, normalizing features by their exponential moving average estimates, \etc. More details can be found in the supplementary material.



\subsection{Generator}
\begin{figure}
\begin{center}
\includegraphics[width=\linewidth]{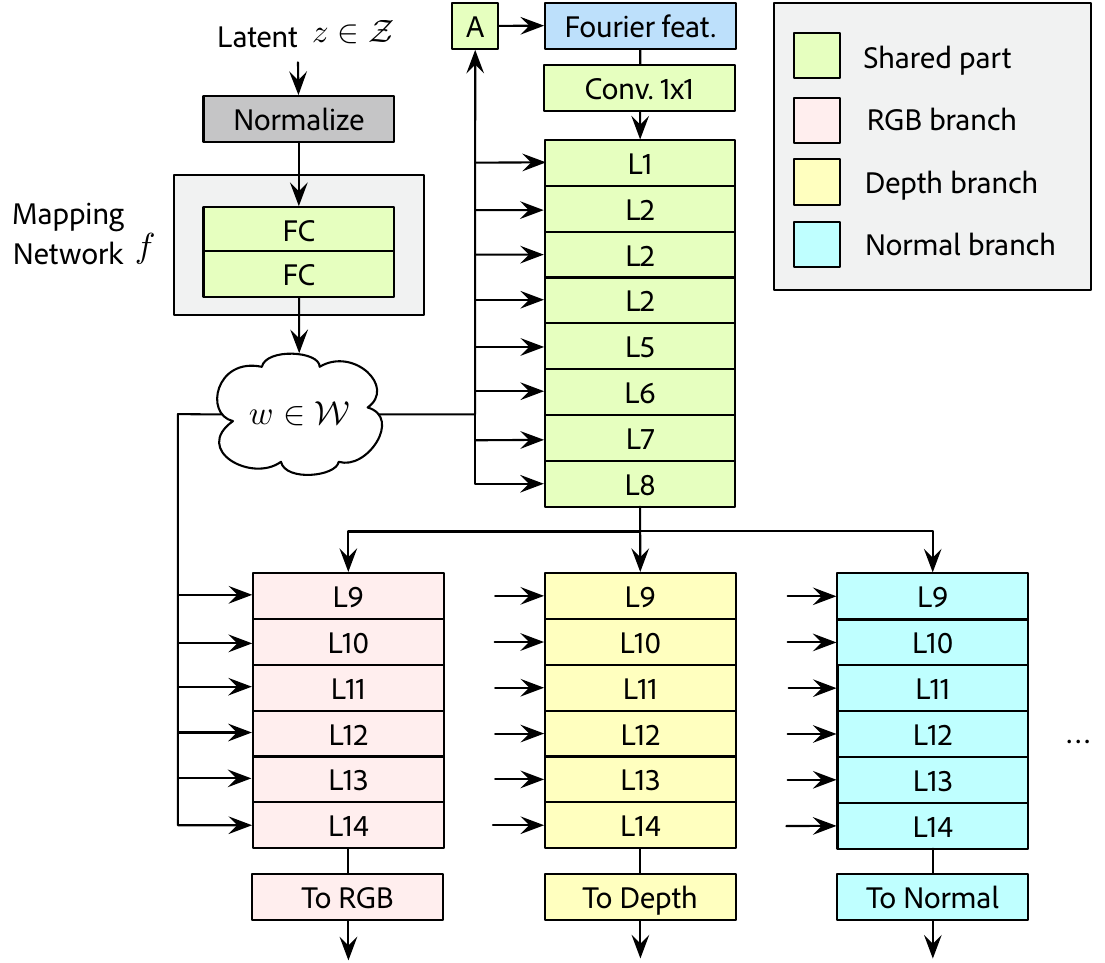}
\end{center}
\vspace{-4mm}
\caption{Diagram of our generator, which employs a shared backbone first and separate branches to generate each modality.}
\label{fig:generator}
\vspace{-4mm}
\end{figure}

The diagram of our generator is depicted in Fig.~\ref{fig:generator}. We keep the same depth as StyleGAN3. The commonly shared part of the generator enumerates until layer 8. Then we design separate branches for the respective modalities. This design considers that different modalities share the same underlying structure of a real scene and we assume that the shared part of our architecture is able to represent the similarities, even without clear guidance. The different branches share the same structure, except for the output layer that uses different numbers of channels for each of the modalities. The outputs of all modalities are grouped as tuples.

\subsection{Discriminators}

We aim to generate data that is realistic and consistent across different modalities. Ideally, this goal should be achieved through only one discriminator that receives the concatenated modalities. 
However, the generator can trivialize the learning of this discriminator by feeding spatially aligned tuples, while some modalities may not look realistic at all. This problem becomes more severe when dealing with more modalities (see Sec.~\ref{sec:experiment}). Therefore, it is necessary to use multiple discriminators that validate the fidelity of each modality separately. However, only using these standalone discriminators also cannot guarantee consistency of the generated tuples.

\begin{figure}
\begin{center}
\includegraphics[width=\linewidth]{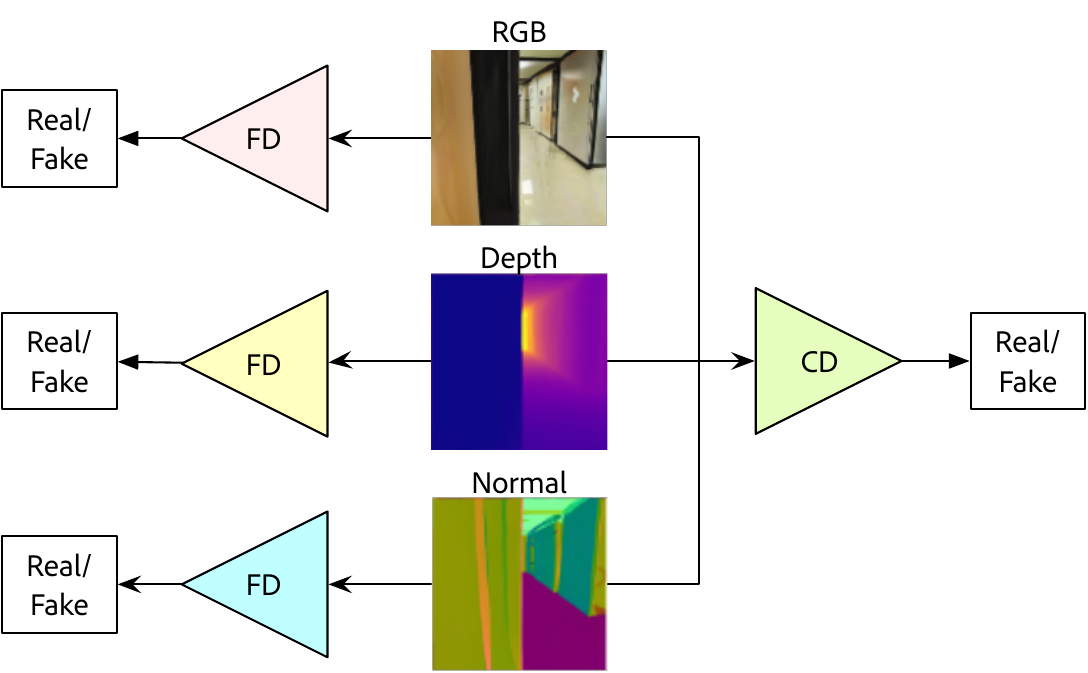}
\end{center}
\vspace{-5mm}
\caption{Diagram of our discriminators. ``CD'' represents consistency discriminator and ``FD'' represents fidelity discriminator of a specific modality.}
\label{fig:discriminator}
\vspace{-4mm}
\end{figure}

Therefore, we developed two types of discriminators. The first type is called \textit{Fidelity Discriminators} (denoted by $D_{F}^m$) used for assessing the fidelity of the corresponding generated modality $m$. The other type of discriminator is called \textit{Consistency Discriminator} (denoted by $D_C$). This discriminator judges whether a tuple of inputs is consistent or not. The input to $D_C$ is the concatenation of all modalities. A diagram of our discriminators is shown in Fig.~\ref{fig:discriminator}. The architecture of each discriminator is generally the same except for different input channels of the first convolutional layers. The structure is inherited from the discriminator design in StyleGAN2~\cite{karras2020styleganv2} and later versions~\cite{ADA,karras2021styleganv3}.

\subsection{Training}
Leveraging supervised training for unconditional generators is usually difficult for a single domain. It becomes feasible in multi-task learning as one can use a synthesized sample of one modality to estimate other modalities, while ensuring consistency. For example, if the generator is trained to produce RGB and depth images simultaneously, we can employ a pre-trained depth estimator~\cite{yin2021depthestimation} to predict depth images from the synthesized RGB images and supervise the synthesized depths images to be closer to estimated ones. However, this approach relies on the performances of the respective models and it may fail when the pre-trained models are trained on a different data distribution. Furthermore, relying on complex exiting architectures increases the compute requirements, which prevents scaling to multiple data modalities. %
Therefore, we mainly use adversarial losses to train the whole system for simplicity and reliability. During training, all discriminators are trained jointly and their outputs are concatenated together:
\begin{equation}
    D_o(x) = [[D_F^m(x^m)], \ D_C(x)],
\end{equation}
where $x^m$ is either generated or real data of a modality $m$ and $x$ is the concatenation of either a generated or real tuple, which means that $x=[x^m]$. 
By default, we use the non-saturating logistic loss~\cite{Ian2014GAN}. Hence, the full objective is:
\begin{equation}
\begin{aligned}
    \min_G \max_D V(D, G)=\mathbb{E}_{x \sim p_{\text {data }}(\mathbf{x})}[\log D_o(\mathbf{x})] \\ -\mathbb{E}_{\mathbf{z} \sim p_{\mathbf{z}}(\mathbf{z})}[\log D_o(G(\mathbf{z}))].
\end{aligned}
\end{equation}

\subsection{Data Augmentation Strategy}
\label{sec:method_data_augmentation}

Collecting data for different tasks is commonly a laborious effort. When performing multimodal generation training, pairwise data is even harder to obtain, which poses great challenges for training our model. In order to cope with data insufficiency, a commonly used technique is data augmentation. As shown in other methods~\cite{diffaugment,ADA}, applying data augmentation operations to real and generated data is beneficial to ensure a balance between discriminator and generator, even when the amount of data is small. We also use data augmentation to train our model, but we face the challenge that it is not possible to use the same augmentation operations across the different data modalities. For example, when generating RGB images and normals jointly, performing a horizontal flip for both modalities does not produce a reliable tuple -- images of surface normals cannot be simply flipped.  

However, as shown in~\cite{ADA}, as long as the discriminator is trained on an adequate number of samples of real data (no augmentations are performed) it can help the generator to approach the real data distribution. Therefore, we apply the same augmentations for all modalities, except for color distortions of RGB images. Even with this seemingly unreasonable augmentation strategy, our results show the consistency of our generated modalities. We apply the same augmentation pipeline as in~\cite{ADA}, since it contains more geometric translations compared to other approaches. We find that these translations are important for modalities capturing geometry and structure. Each augmentation operation in the pipeline has the same execution probability $p$ of being executed or $1-p$ of not being executed. We follow the same approach as described in their paper to adjust $p$. Moreover, we treat $p$ as a signal to stop training when it reaches a threshold, i.e., 0.7 in our case, since larger values of $p$ also indicates severe overfitting of the discriminator. When $p$ exceeds the threshold, it also becomes increasingly rare for the generator to see true training samples, thus causing generation with distortions.

\vspace{-1mm}
\subsection{Cross-modal Fine-tuning}
\label{sec:cross_model_fine_tuning}
\vspace{-1mm}


In certain domains, there are often abundant RGB images yet scarce corresponding labels (e.g., depth). For instance, a construction site may possess ample RGBD pairs for most rooms (domain A), but insufficient pairs for a specific scene, such as a utility room (domain B).
This data imbalance makes training a generator from scratch challenging. To address this, we employ cross-domain fine-tuning, wherein a model pretrained on domain A is fine-tuned to accommodate domain B, despite limited pairwise data.

In our framework, only the consistency discriminator necessitates pairwise supervision, and therefore, is exclusively trained on pairwise data. Conversely, all available data can be used to train the fidelity discriminators. However, the limited pairwise data may skew the training of different discriminators, empirically leading to inconsistent generation or even optimization divergence. To rectify this imbalance, we modify the data sampling procedure to sample paired and unpaired data equally. The paired data provides accurate annotations that assist the generator in generalization, while the unpaired data introduces diversity to the corresponding modalities. Through the comprehensive use of all data, we enable successful generalization to a novel domain.


%

\section{Experiments}
\label{sec:experiment}

\subsection{Dataset}
We focus on the Stanford2D3D~\cite{armeni2017stanford2d3d} dataset for evaluation. It contains 70,496 pairwise inputs of multiple data modalities, such as RGB, depth, surface normal, etc. Images in this dataset contain different types of complex scenes with various objects (e.g., chair, shelves, monitor) inside a Stanford building, rather than a single object located at the center of an image. 
Previous works~\cite{shi2022cuhk} mainly experimented with LSUN-Bedroom and LSUN-Kitchen~\cite{LSUN} and show that networks based on StyleGAN can generate realistic images on these datasets, while ours is the first to show the capacity of StyleGAN generators to deal with such a complex dataset. Please refer to the supplementary material for implementation details.

\subsection{Evaluation Metrics}
Since our model is unconditional, it is not feasible to directly compare generated and real samples in a pairwise manner. Instead we adopt FID~\cite{FID} to calculate the distance between distributions of generated data and real data. We feed data of a particular modality to an InceptionV3~\cite{InceptionV3} model pre-trained on ImageNet~\cite{ImageNet} to obtain feature means and variances. Then, we calculate the Frechet distance using these statistics averaged over a fraction of sampled data points. For real data, we sample $\min (50\mathrm{K}, |\mathcal{D}_{\mathrm{train}}|)$ real data points, where $|\mathcal{D}_{\mathrm{train}}|$ indicates the number of data points in the training dataset. For calculating the statistics in generated data, we generate 50K data points.

Besides assessing the distribution discrepancy, we also want to ensure that the different modalities are consistent with each other. In order to evaluate consistency, we first generate 5K pairwise samples. For depth, we calculate the averaged Scale Invariant Depth Error (\textbf{SIE})~\cite{SIE} between a generated depth map and an estimated depth map, obtained by feeding the generated RGB to a pre-trained depth estimator~\cite{yin2021depthestimation}. For surface normals, we use an off-the-shelf surface normal estimator~\cite{eftekhar2021omnidata,kar20223d} to estimate surface normals from a generated RGB image and then compare the result with our generated normals. To assess normal consistency with generated RGB images, we use the standard metrics proposed in prior work~\cite{WangFG15}: the mean and median of angular error measured in degrees, and the percentage of pixels whose angular error is within $\gamma$ degrees.

\subsection{Comparison}
\begin{figure*}[t]
\begin{center}
\includegraphics[width=\linewidth]{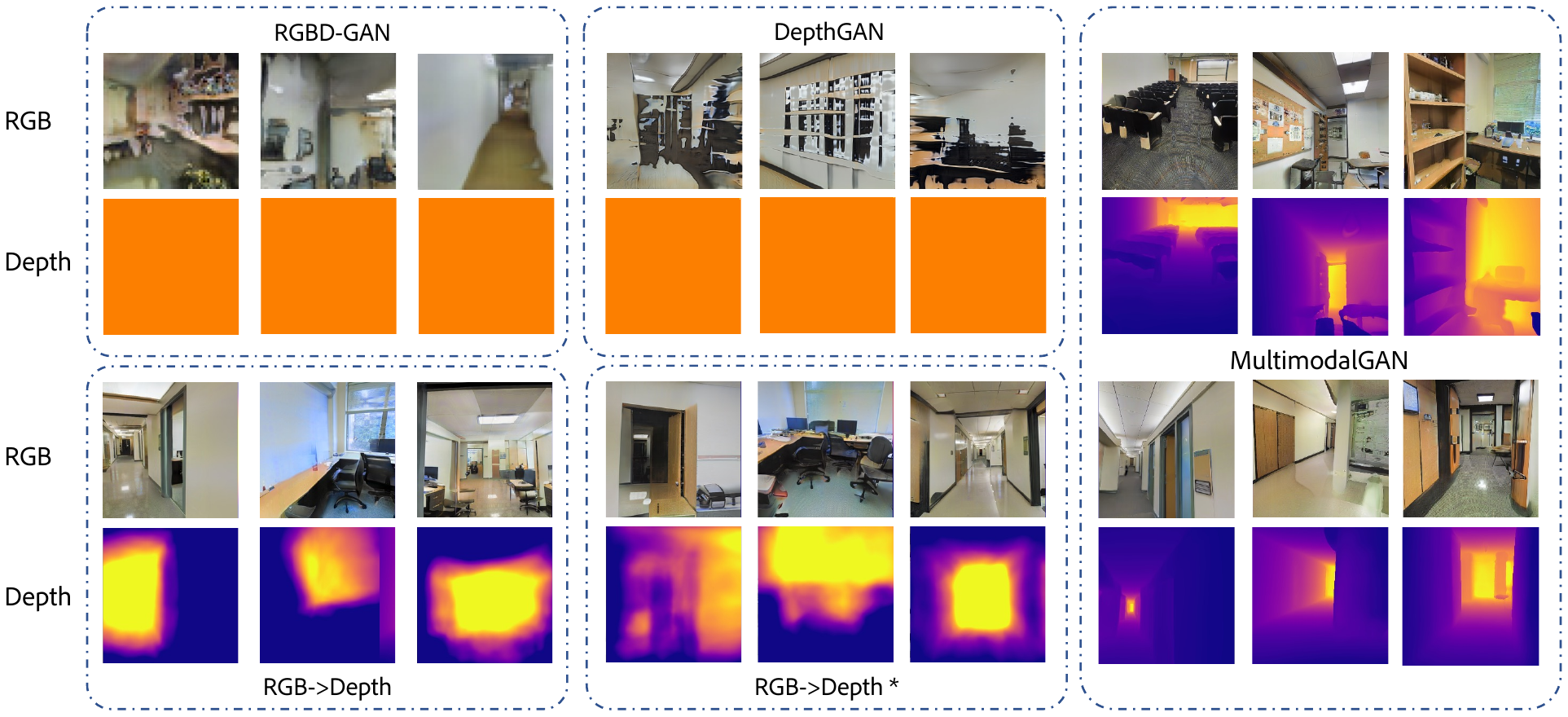}
\end{center}
\vspace{-6mm}
\caption{Qualitative comparison between the proposed MultimodalGAN and previous methods. All generated depth maps are visualized using the same color map.}
\vspace{-4mm}
\label{fig:comparison}
\end{figure*}

We compare with methods that address unconditional synthesis of multiple modalities. To this end, the DepthGAN~\cite{shi2022cuhk} and RGBD-GAN~\cite{noguchi2019rgbd} are most closely related to our work. We use their original codebases to run experiments on Stanford2D3D dataset and keep their original model and settings for a fair comparison. DatasetGAN~\cite{zhang2021datasetgan} is based on a StyleGAN generator for learning additional tasks, such as segmentation and key-point estimation. The underlying assumption is that the StyleGAN generator already provides a geometrical or disentangled understanding of the generated image. Therefore, the only effort is to extract labels from intermediate representations of the generator; a few images need to be generated and then manually annotated. Finally, they trained a small network on top of StyleGAN generator to translate features into dense labels. However, the efficacy of DatasetGAN is only demonstrated for a single-object dataset. 
A limitation of this approach is that some modalities cannot be annotated by human annotators -- for example, depth is captured instead of annotated. Therefore, we cannot directly compare with DatasetGAN. To validate our model, we first train a depth estimation model~\cite{yin2021depthestimation} using all real RGB and depth pairs. Then we pre-train a standard StyleGAN3~\cite{karras2021styleganv3} model on RGB images. After that, we build a depth branch on top of the pre-trained generator, which results in an identical architecture to our generator with RGB and depth branches. We fix all parameters except for those in the depth branch and leverage supervised training to train the depth branch by treating the estimated depth from the LeReS model on generated RGB image as ground-truth. We denote this approach as ``\rgbd''. LeReS~\cite{yin2021depthestimation} also provides well-trained models on lots of data. We also use these pre-trained representations to provide guidance for training the depth branch and compare the performance with our approach (\rgbd*).

In Tab.~\ref{table:comparison} we show that our method achieves good RGB-FID and the best Depth-FID and SIE scores, demonstrating the effectiveness of our approach. Both the {\rgbd} model and {\rgbd*} model have very large Depth-FIDs and SIEs. 
%
%
\zzhu{The reason lies in the incompatibility between the inputs to the depth branch and depth modality: either the feature maps from L8 or the intermediate latent codes $w$ are optimized towards the RGB modality, posing large challenge when learned towards depth, especially other parts are fixed except for the depth branch (See~\cref{fig:generator}). This further confirms the benefit of learning multiple modalities jointly.}
In Fig.~\ref{fig:comparison}, both RGBD-GAN and DepthGAN fail to generate plausible depth, while RGBD-GAN is comparatively better at RGB images. {\rgbd} and {\rgbd*} produce better depth results, but are still far from satisfaction compared to ours. This consolidates the effectiveness of our approach in generating multiple modalities with one model.


\begin{table}[t]
  \caption{Quantitative comparisons with other methods. For all metrics, the lower, the better.}
  \label{table:comparison}
  \centering
  \scalebox{0.93}{
  \begin{tabular}{l|ccc}
    \toprule
    Method & RGB-FID & Depth-FID & SIE \\
    \midrule
    RGBD-GAN~\cite{noguchi2019rgbd}  & 105.0 & 46.7 & 0.339  \\
    DepthGAN~\cite{shi2022cuhk}  & 326.3 & 182.9 & 0.467  \\
    \rgbd  & 13.8 & 227.5 & 1.745  \\
    \rgbd*  & 13.8 & 245.9 & 1.182 \\
    \textbf{MultimodalGAN}  & 14.6 & 24.8 & 0.192  \\
    \bottomrule
  \end{tabular}}
\end{table}

\subsection{Ablation}

To further explore the efficacy of our approach we provide a set of ablation studies. Most of the ablation experiments are conducted on the RGB and depth modalities.

\begin{table}[t]
	\centering
	\caption{Ablation in terms of data augmentation strategies.}
	\vspace{-2mm}
	\scalebox{0.93}{
	\begin{tabular}{ll|ccc}
		\toprule
		Model & Aug. & RGB-FID & Depth-FID & SIE \\
		\midrule
		RGB & No aug. & 41.6 & - & - \\ 
		RGB & Diff.~\cite{diffaugment} & 29.2 & - & - \\ 
	    RGB & ADA~\cite{ADA} & 12.6 & - & - \\
		Full & No aug. & 35.3 & 33.3 & 0.236 \\ 
		Full & Diff.~\cite{diffaugment} & 20.6 & 29.9 & 0.193 \\ 
	    Full & ADA~\cite{ADA} & 14.6 & 24.8 & 0.192 \\
		\bottomrule
	\end{tabular}}
	\vspace{-4mm}
	\label{table:ablation_augmentation}
\end{table}
\vspace{0.5em}\noindent\textit{\textbf{Choice of augmentation strategies.}}~
As discussed in Sec.~\ref{sec:method_data_augmentation}, data augmentation is vital, especially when the amount of data is not sufficient for effectively training a modoel.
Multiple discriminator augmentation strategies can be used to improve performance of GAN training, including DiffAugment~\cite{diffaugment} and ADA~\cite{ADA}. For a comparison of different augmentation methods, we use two models: 1) RGB baseline: we directly train a StyleGAN3 model on RGB images from the Stanford2D3D dataset; 2) Full model: our final model as shown in Fig.~\ref{fig:generator} and Fig.~\ref{fig:discriminator}. For the full model, we apply the same augmentation strategy to both modalities. As shown in Tab.~\ref{table:ablation_augmentation}, ADA provides the best results for both models. Our hypothesis is that the performance improvement is mainly caused by geometric translations in ADA. In our case, the depth modality requires modelling scene structures, which can be enhanced by geometric translation operations. As shown in~\cite{ADA}, geometric translations are helpful when the amount of data is small.

\begin{table}[t]
	\centering
	\caption{Ablation in terms of generator architecture on Stanford2D3D dataset. }
	\vspace{-2mm}
	\scalebox{0.9}{
	\begin{tabular}{l|ccc}
		\toprule
		Method & RGB-FID & Depth-FID & SIE \\
		\midrule
		RGB baseline & 12.6 & - & - \\ 
		Depth baseline & - & 20.9 & - \\ 
		RGBD baseline & 15.5 & 26.2 & 0.189 \\
		Branch baseline & 13.6 & 22.9 & 0.192  \\ 
		\bottomrule
	\end{tabular}}
	\label{table:ablation_baselines}
	\vspace{-4mm}
\end{table}
\vspace{0.5em}\noindent\textit{\textbf{Building separate branches for different modalities.}}~
Our generator contains a shared part and then branches out for the respective data modalities. Alternatively, one can build only one branch with the final output layer to produce multi-channel outputs, which can be split to represent different modalities. We refer to this baseline as {\bf RGBD baseline} when producing RGB and depth simultaneously. This baseline can be trained with one discriminator which receives a four-channel input (first three channels for RGB, the rest one for depth). As another baseline, we also train a StyleGAN3 model on the RGB and depth domains, which we refer to as {\bf RGB baseline} and {\bf Depth baseline}. These models are all trained with one discriminator. For fair comparison, we build a branch generator with only one discriminator, called {\bf Branch baseline}, for which we build separate branches from L9 as shown in Fig.~\ref{fig:generator}.

The quantitative comparison is shown in Tab.~\ref{table:ablation_baselines}. For the multi-modality baselines, both the RGB-FIDs and Depth-FIDs are lower than those of the single-modality baselines. However, the RGB baseline and the Depth baseline alone do not have many practical applications compared to the multimodal generation models. When comparing the RGBD baseline with the Branch baseline, we observe better results for the Branch baseline in both FID metrics (with a slightly inferior SIE). This shows the advantage of generating data with better quality and decent consistency.


\begin{table}[t]
	\centering
	\caption{Ablation in terms of the layer indexes to build branches.}
	\vspace{-2mm}
	\scalebox{0.9}{
	\begin{tabular}{l|ccc}
		\toprule
		Layer index & RGB-FID & Depth-FID & SIE \\
		\midrule
		L6 & 13.3 & 23.5 & 0.207 \\ 
		L9 & 14.6 & 24.8 & 0.192\\ 
		L12 & 15.3 & 25.6 & 0.191 \\
		\bottomrule
	\end{tabular}}
	\label{table:ablation_branch_layer}
\vspace{-1em}
\end{table}
\vspace{0.5em}\noindent\textit{\textbf{Layer indexes to build branches.}}~
After the generator architecture, we investigate the correct place to build branches in the generator. The RGBD baseline can be treated as an extreme case when regarding the branching index as L13. So we further investigate smaller indices. We experimented with defining branching from L6, L9 and L12 along with the different spatial sizes of their output features. 

As illustrated in Tab.~\ref{table:ablation_branch_layer}, if the starting branch is L6, both FID measurements are comparatively inferior to others, since this model fails to converge. Our hypothesis is that the formation of shared underlying structures is not complete in the shallow layers. As training starts, the outputs of different modalities do not align well, causing the consistency discriminator to easily differentiate real/fake pairs which consequently leads to unsuccessful training. Since the generation quality of RGB is not satisfactory, the SIE metric is high. Using L9 already provides reasonable results and we observe that the qualitative results from L12 show more higher frequency artifacts (the depth modality contains high frequency noise). Therefore, we choose L9 to build branches as default.

\begin{table*}[htbp]
	\centering
	\caption{Ablation in terms of discriminator architecture. \emph{w.} normal represents whether to add the normal modality into the generation task.}
	\vspace{-2mm}
	\scalebox{0.9}{
	\begin{tabular}{lc|c|cc|cccccc}
		\toprule
		 & & RGB & \multicolumn{2}{c|}{Depth} & \multicolumn{6}{|c}{Surface Normal} \\
		 \midrule
		\multirow{2}{*}{Disc.} & \multirow{2}{*}{\emph{w.} normal} & \multirow{2}{*}{RGB-FID} & \multirow{2}{*}{Depth-FID} & \multirow{2}{*}{SIE} & \multirow{2}{*}{Normal-FID} & \multicolumn{2}{c}{Anglular Error$^{\circ}$ $\downarrow$} & \multicolumn{3}{c}{$\%$ Within $\gamma^{\circ}$ $\uparrow$}\\
		 & & & & & & Mean & Median & 11.25$^{\circ}$ & 22.5$^{\circ}$ & 30$^{\circ}$ \\
		\midrule
		FD &  & 14.1 & 25.3 & 0.298 & - & - & - & - & - & - \\ 
		CD & & 13.6 & 22.9 & 0.192 & - & - & - & - & - & - \\ 
		CD+FD &  & 14.6 & 24.8 & 0.192 & - & - & - & - & - & - \\
		CD & $\checkmark$ & 185.7 & 48.5 & 0.360 & 44.0 & 20.36 & 17.23 & 32.24 & 62.73 & 77.39 \\ 
		CD+FD & $\checkmark$  & 15.6 & 29.0 & 0.195 & 15.4 & 14.77 & 12.23 & 46.40 & 79.69 & 89.87 \\
		\bottomrule
	\end{tabular}}
	\label{table:ablation_discriminator}
\vspace{-1em}
\end{table*}
\vspace{0.5em}\noindent\textit{\textbf{Discriminator design.}}~
In our full model, we use the fidelity discriminators (FD) and the consistency discriminator (CD) to ensure both generation quality and modality consistency. We aim to show the effect of each type to the performance by removing either type from the Full model. In Tab.~\ref{table:ablation_discriminator}, we show that when the model only has two modalities, using only the consistency discriminator would suffice. However, when we scale up to more modalities (adding surface normals), using both types of discriminators provide the best performance. Using only the fidelity discriminators cannot guarantee a consistent tuple across modalities, as indicated by the large SIE.

\vspace{0.5em}\noindent\textit{\textbf{Scale up to more modalities.}}~
Our model is not constrained to only two modalities. As shown in Tab.~\ref{table:ablation_discriminator}, we can further scale up to also generate surface normals. Training with three modalities can be accomplished without a significant drop in performance for the modalities, when using both consistency discriminator and fidelity discriminators. As we can see from the result in Fig.~\ref{fig:teaser}, the results generated by this model show diversity over many scene classes of Standford2D3D dataset and complex layouts.

\section{Applications}

\begin{figure}
\begin{center}
\includegraphics[width=\linewidth]{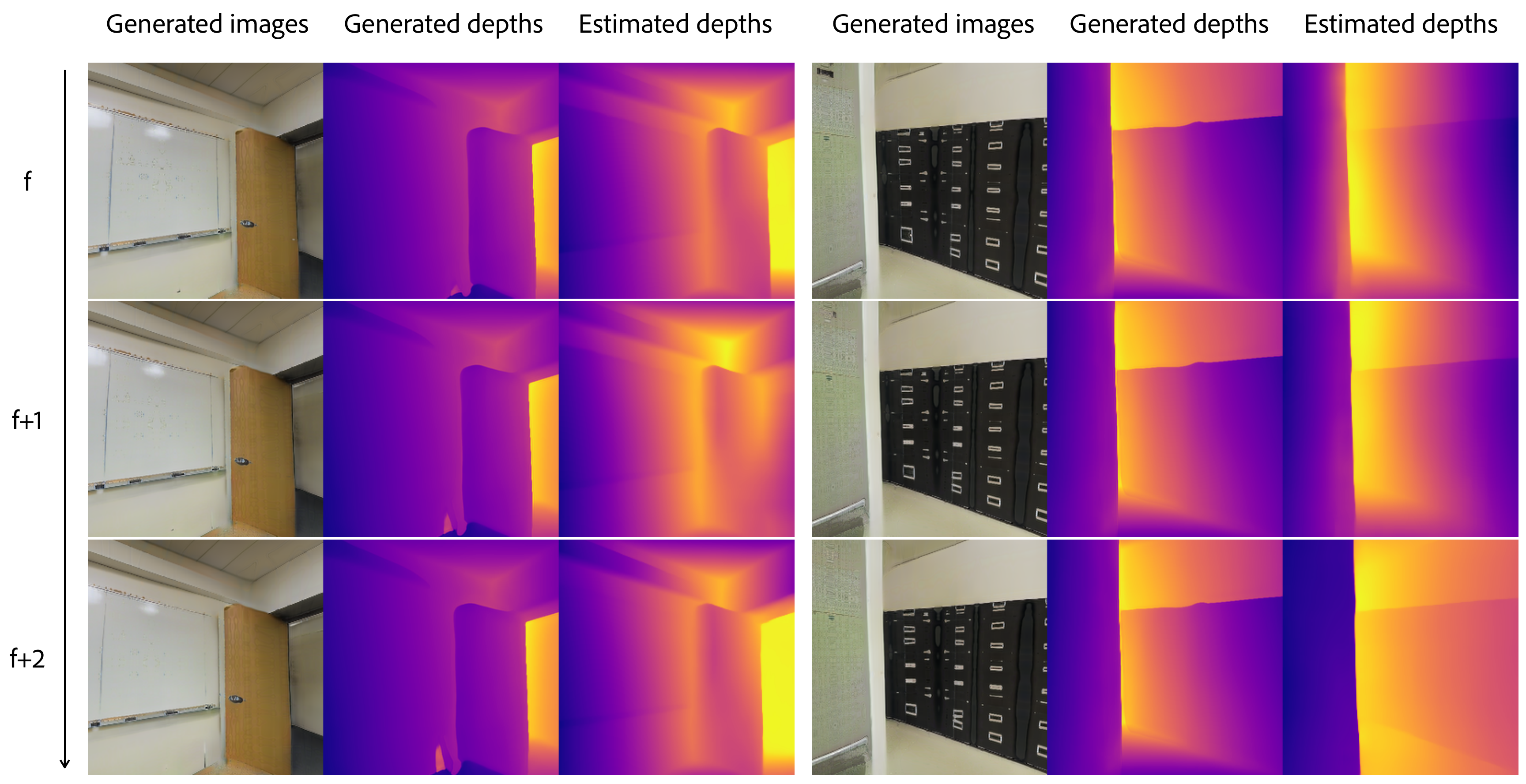}
\end{center}
\vspace{-6mm}
\caption{Qualitative results showing the smoothness of generated depths versus estimated depths from ~\cite{yin2021depthestimation}. ``f'' represents the the initial frame index.}
\label{fig:smooth_depth}
\vspace{-5mm}
\end{figure}
\subsection{Consistent RGBD Generation}

\zzhu{One advantage of StyleGAN-based approaches is their capacity to generate smooth outputs through linear interpolation in the latent space. Our model retains this capability across not only the conventional RGB modality but also additional modalities. This is an obvious advantage of our approach over existing methods that estimate corresponding modalities from each RGB image, as these estimations can exhibit considerable flickering issue in subsequent interpolations. As demonstrated in~\cref{fig:smooth_depth}, our generated depths maintain consistent layouts and local details across all time frames, whereas the estimated depths show substantial inconsistencies in extensive regions in successive frames. 
More video results can be found in the supplementary materials.}

\begin{table}[t]
	\centering
	\caption{Fine-tuning on held-out auditorium scene using different portions of pairwise data. The first two columns represent the portion of images used for training.}
	\scalebox{0.9}{
	\begin{tabular}{ll|ccc}
		\toprule
		RGB+D & RGB &  RGB-FID & Depth-FID & SIE \\
		\midrule
		5\% & 95\% & 66.3 & 115.4 & 0.164 \\ 
		10\% & 90\% & 53.8 & 102.3 & 0.158 \\ 
		15\% & 85\% & 53.0 & 91.3 & 0.144 \\
		\bottomrule
	\end{tabular}}
	\label{table:finetuning}
\vspace{-4mm}
\end{table}

\subsection{Cross-domain Fine-tuning}

\zzhu{To demonstrate cross-domain fine-tuning, we partition the Stanford2D3D dataset into two subsets, designating the ``auditorium'' scene (comprising 6,371 pairwise data) as the target domain, and the remaining scene classes as the source domain. In the target domain, we employ only a percentage $\mathbf{p}\%$ of the pairwise data, assuming the model has access to the remaining $(1-\mathbf{p}\%)$ RGB images. The source domain undergoes pre-training to concurrently generate RGB images and depth. Upon completion, we apply the fine-tuning procedure outlined in~\cref{sec:cross_model_fine_tuning} to the pre-trained model on the target domain. We select $\mathbf{p}$ from the set ${\mathrm{5, 10, 15}}$ to exhibit the model's performance with varying amounts of pairwise data for fine-tuning. As illustrated in Tab.~\ref{table:finetuning}, all metrics decrease with the incorporation of more pairwise data. According to the RGB-FID, 10\% of RGBD pairs suffice for fine-tuning. We also present qualitative results and corresponding videos in the supplementary material, showing the model's satisfactory generalization to a new scene, exhibiting considerable latent smoothness and quality, even with minimal fine-tuning.}

%
%

\subsection{Generating Datasets for Depth Estimation}
\label{sec:generating_dataset}
\vspace{-1mm}

\begin{figure}
\begin{center}
\includegraphics[width=\linewidth]{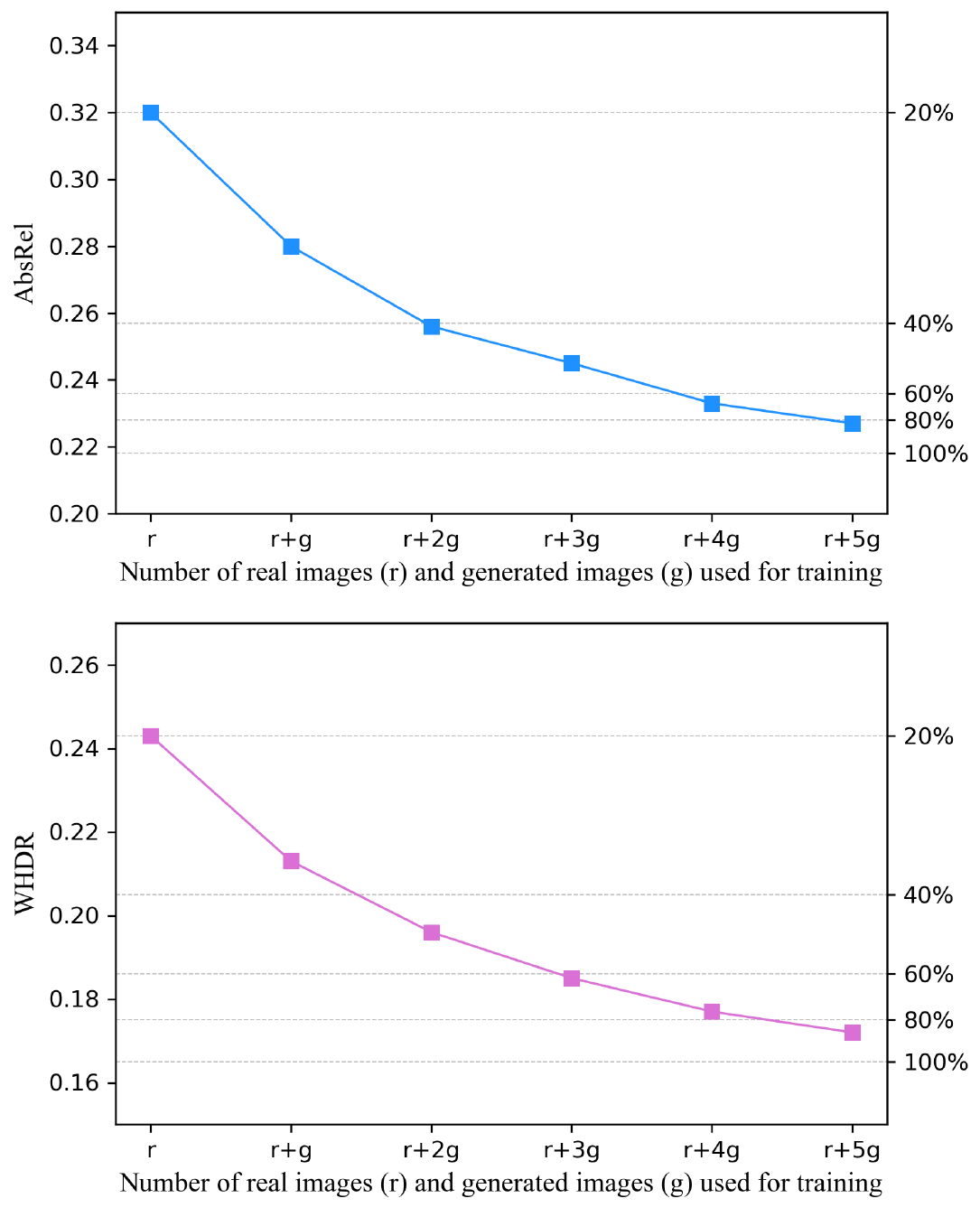}
\end{center}
\vspace{-6mm}
\caption{Depth estimation performance of LeReS~\cite{yin2021depthestimation} when trained on different combinations of real data and generated data. The $x$-axis represents the combination of real and generated images. The left $y$-axes are the numbers of AbsRel or WHDR obtained by the corresponding models, respectively. The dashed horizontal lines indicate the performances of training on different portions of training dataset. ``$r$''/ ``$g$'' represent the numbers of real/generated images. In this case, $r=g=12,045$, which is 20\% size of the training set.}
\label{fig:depth_estimation}
\vspace{-5mm}
\end{figure}

Can we generate a synthetic dataset to benefit downstream tasks?  
To explore 
this objective, we synthesize a dataset of RGB and depth images with our multi-modal GAN and validate its usage in the task of depth estimation.

We first split the Stanford2D3D dataset into training and validation sets using a ratio of 6:1. Of the training samples, 20\% (12,045 RGB+D pairs) are randomly chosen to train our generative model. Despite the small training set, the baseline achieves good performance:  the RGB-FID, Depth-FID and SIE of this model are 23.7, 27.0 and 0.192, respectively.
Then, we use the model to produce a series of synthetic datasets, of one to five times 
the initial 20\% set.  
%
We train LeReS~\cite{yin2021depthestimation} from scratch on each of these datasets for 30K iterations, since the models converge well within this amount of training. We measure performance  with absolute mean relative error (AbsRel) and Weighted Human Disagreement Rate (WHDR)~\cite{WHDR} for evaluation. More details are provided in the supplementary material. As shown in Fig.~\ref{fig:depth_estimation}, using more of our generated data consistently improves both metrics.  When the total number of generated and real samples is equal to the size of the full training set ($r+4g$ on the $x$-axis), the model achieves similar performance to using 80\% of the real data. This is impressive, given that the generator is only trained with 20\% of real data, and indicates that our generative model can effectively generate synthetic training data. 



\section{Conclusion}
\zzhu{We introduce a novel architecture to generate multiple image modalities simultaneously. We demonstrate its effectivenss on the challenging Stanford2D3D dataset and several interesting applications. Especially, our architecture can predict the modalities coherently, enabling synthesis of datasets that, in turn, can be employed to enhance the performance of downstream tasks. Currently, we attempted generating RGB, depth and surface normal under the GAN framework. In the future, we would like to try more modalities such as segmentation mask and other generative models like diffusion models. 
}

{\small
\bibliographystyle{ieee_fullname}
\bibliography{egbib}

\begin{thebibliography}{10}\itemsep=-1pt

\bibitem{armeni2017stanford2d3d}
Iro Armeni, Sasha Sax, Amir~R Zamir, and Silvio Savarese.
\newblock Joint 2d-3d-semantic data for indoor scene understanding.
\newblock {\em arXiv preprint arXiv:1702.01105}, 2017.

\bibitem{bao2022generative}
Zhipeng Bao, Martial Hebert, and Yu-Xiong Wang.
\newblock Generative modeling for multi-task visual learning.
\newblock In {\em Proc. ICML}, 2022.

\bibitem{brock2018large}
Andrew Brock, Jeff Donahue, and Karen Simonyan.
\newblock Large scale gan training for high fidelity natural image synthesis.
\newblock {\em arXiv preprint arXiv:1809.11096}, 2018.

\bibitem{7989023}
Arunkumar Byravan and Dieter Fox.
\newblock Se3-nets: Learning rigid body motion using deep neural networks.
\newblock 2017.

\bibitem{casser2019struct2depth}
Vincent Casser, S\"{o}ren Pirk, Reza Mahjourian, and Anelia Angelova.
\newblock Depth prediction without the sensors: Leveraging structure for
  unsupervised learning from monocular videos.
\newblock In {\em AAAI}, 2019.

\bibitem{crawshaw2020surveymtl}
Michael Crawshaw.
\newblock Multi-task learning with deep neural networks: A survey.
\newblock {\em arXiv preprint arXiv:2009.09796}, 2020.

\bibitem{dhariwal2021diffusion}
Prafulla Dhariwal and Alexander Nichol.
\newblock Diffusion models beat gans on image synthesis.
\newblock In {\em NeurIPS}, 2021.

\bibitem{eftekhar2021omnidata}
Ainaz Eftekhar, Alexander Sax, Jitendra Malik, and Amir Zamir.
\newblock Omnidata: A scalable pipeline for making multi-task mid-level vision
  datasets from 3d scans.
\newblock In {\em CVPR}, 2021.

\bibitem{SIE}
David Eigen, Christian Puhrsch, and Rob Fergus.
\newblock Depth map prediction from a single image using a multi-scale deep
  network.
\newblock In {\em NeurIPS}, 2014.

\bibitem{Ian2014GAN}
Ian Goodfellow, Jean Pouget-Abadie, Mehdi Mirza, Bing Xu, David Warde-Farley,
  Sherjil Ozair, Aaron Courville, and Yoshua Bengio.
\newblock Generative adversarial nets.
\newblock In {\em NeurIPS}, 2014.

\bibitem{FID}
Martin Heusel, Hubert Ramsauer, Thomas Unterthiner, Bernhard Nessler, and Sepp
  Hochreiter.
\newblock Gans trained by a two time-scale update rule converge to a local nash
  equilibrium.
\newblock In {\em NeurIPS}, 2017.

\bibitem{kar20223d}
O{\u{g}}uzhan~Fatih Kar, Teresa Yeo, Andrei Atanov, and Amir Zamir.
\newblock 3d common corruptions and data augmentation.
\newblock In {\em CVPR}, 2022.

\bibitem{progressivegan}
Tero Karras, Timo Aila, Samuli Laine, and Jaakko Lehtinen.
\newblock Progressive growing of gans for improved quality, stability, and
  variation.
\newblock In {\em ICLR}, 2018.

\bibitem{ADA}
Tero Karras, Miika Aittala, Janne Hellsten, Samuli Laine, Jaakko Lehtinen, and
  Timo Aila.
\newblock Training generative adversarial networks with limited data.
\newblock In {\em NeurIPS}, 2020.

\bibitem{karras2021styleganv3}
Tero Karras, Miika Aittala, Samuli Laine, Erik H{\"a}rk{\"o}nen, Janne
  Hellsten, Jaakko Lehtinen, and Timo Aila.
\newblock Alias-free generative adversarial networks.
\newblock 2021.

\bibitem{karras2019styleganv1}
Tero Karras, Samuli Laine, and Timo Aila.
\newblock A style-based generator architecture for generative adversarial
  networks.
\newblock In {\em CVPR}, 2019.

\bibitem{karras2020styleganv2}
Tero Karras, Samuli Laine, Miika Aittala, Janne Hellsten, Jaakko Lehtinen, and
  Timo Aila.
\newblock Analyzing and improving the image quality of stylegan.
\newblock In {\em CVPR}, 2020.

\bibitem{PolymorphicGAN}
Seung~Wook Kim, Karsten Kreis, Daiqing Li, Antonio Torralba, and Sanja Fidler.
\newblock Polymorphic-gan: Generating aligned samples across multiple domains
  with learned morph maps.
\newblock In {\em CVPR}, pages 10620--10630, 2022.

\bibitem{adam}
Diederik~P. Kingma and Jimmy Ba.
\newblock Adam: {A} method for stochastic optimization.
\newblock In Yoshua Bengio and Yann LeCun, editors, {\em ICLR}, 2015.

\bibitem{kingma2013auto}
Diederik~P Kingma and Max Welling.
\newblock Auto-encoding variational bayes.
\newblock {\em arXiv preprint arXiv:1312.6114}, 2013.

\bibitem{semanticGAN}
Daiqing Li, Junlin Yang, Karsten Kreis, Antonio Torralba, and Sanja Fidler.
\newblock Semantic segmentation with generative models: Semi-supervised
  learning and strong out-of-domain generalization.
\newblock In {\em CVPR}, pages 8300--8311, 2021.

\bibitem{lu2021taskology}
Yao Lu, S\"{o}ren Pirk, Jan Dlabal, Anthony Brohan, Ankita Pasad, Zhao Chen,
  Vincent Casser, Anelia Angelova, and Ariel Gordon.
\newblock Taskology: Utilizing task relations at scale.
\newblock In {\em CVPR}, 2021.

\bibitem{r1regularization}
Lars~M. Mescheder, Andreas Geiger, and Sebastian Nowozin.
\newblock Which training methods for gans do actually converge?
\newblock 2018.

\bibitem{nyud}
Pushmeet~Kohli Nathan~Silberman, Derek~Hoiem and Rob Fergus.
\newblock Indoor segmentation and support inference from rgbd images.
\newblock In {\em ECCV}, 2012.

\bibitem{noguchi2019rgbd}
Atsuhiro Noguchi and Tatsuya Harada.
\newblock Rgbd-gan: Unsupervised 3d representation learning from natural image
  datasets via rgbd image synthesis.
\newblock {\em arXiv preprint arXiv:1909.12573}, 2019.

\bibitem{ImageNet}
Olga Russakovsky, Jia Deng, Hao Su, Jonathan Krause, Sanjeev Satheesh, Sean Ma,
  Zhiheng Huang, Andrej Karpathy, Aditya Khosla, Michael~S. Bernstein,
  Alexander~C. Berg, and Li Fei{-}Fei.
\newblock Imagenet large scale visual recognition challenge.
\newblock {\em IJCV}, 115(3):211--252, 2015.

\bibitem{shi2022cuhk}
Zifan Shi, Yujun Shen, Jiapeng Zhu, Dit-Yan Yeung, and Qifeng Chen.
\newblock 3d-aware indoor scene synthesis with depth priors.
\newblock {\em arXiv preprint arXiv:2202.08553}, 2022.

\bibitem{InceptionV3}
Christian Szegedy, Vincent Vanhoucke, Sergey Ioffe, Jonathon Shlens, and
  Zbigniew Wojna.
\newblock Rethinking the inception architecture for computer vision.
\newblock In {\em CVPR}, 2016.

\bibitem{WangFG15}
Xiaolong Wang, David~F. Fouhey, and Abhinav Gupta.
\newblock Designing deep networks for surface normal estimation.
\newblock In {\em CVPR}, 2015.

\bibitem{wang2016ssgan}
Xiaolong Wang and Abhinav Gupta.
\newblock Generative image modeling using style and structure adversarial
  networks.
\newblock In {\em ECCV}, 2016.

\bibitem{WHDR}
Ke Xian, Chunhua Shen, Zhiguo Cao, Hao Lu, Yang Xiao, Ruibo Li, and Zhenbo Luo.
\newblock Monocular relative depth perception with web stereo data supervision.
\newblock In {\em CVPR}, 2018.

\bibitem{yin2021depthestimation}
Wei Yin, Jianming Zhang, Oliver Wang, Simon Niklaus, Long Mai, Simon Chen, and
  Chunhua Shen.
\newblock Learning to recover 3d scene shape from a single image.
\newblock In {\em CVPR}, 2021.

\bibitem{LSUN}
Fisher Yu, Yinda Zhang, Shuran Song, Ari Seff, and Jianxiong Xiao.
\newblock {LSUN:} construction of a large-scale image dataset using deep
  learning with humans in the loop.
\newblock {\em CoRR}, abs/1506.03365, 2015.

\bibitem{9156702}
Amir~R. Zamir, Alexander Sax, Nikhil Cheerla, Rohan Suri, Zhangjie Cao,
  Jitendra Malik, and Leonidas~J. Guibas.
\newblock Robust learning through cross-task consistency.
\newblock In {\em CVPR}, 2020.

\bibitem{zamir2018taskonomy}
Amir~R Zamir, Alexander Sax, William Shen, Leonidas~J Guibas, Jitendra Malik,
  and Silvio Savarese.
\newblock Taskonomy: Disentangling task transfer learning.
\newblock In {\em CVPR}, 2018.

\bibitem{zhang2021datasetgan}
Yuxuan Zhang, Huan Ling, Jun Gao, Kangxue Yin, Jean-Francois Lafleche, Adela
  Barriuso, Antonio Torralba, and Sanja Fidler.
\newblock Datasetgan: Efficient labeled data factory with minimal human effort.
\newblock In {\em CVPR}, 2021.

\bibitem{zhang2021surveymtl}
Yu Zhang and Qiang Yang.
\newblock A survey on multi-task learning.
\newblock {\em IEEE Transactions on Knowledge and Data Engineering}, 2021.

\bibitem{diffaugment}
Shengyu Zhao, Zhijian Liu, Ji Lin, Jun{-}Yan Zhu, and Song Han.
\newblock Differentiable augmentation for data-efficient {GAN} training.
\newblock In {\em NeurIPS}, 2020.

\bibitem{8100183}
Tinghui Zhou, Matthew Brown, Noah Snavely, and David~G. Lowe.
\newblock Unsupervised learning of depth and ego-motion from video.
\newblock In {\em CVPR}, 2017.

\end{thebibliography}
}

\clearpage

\appendix
\section{More Details About Method}
While StyleGAN3~\cite{karras2021styleganv3} uses critical sampling in the last two layers of the generator to ensure a good balance between antialiasing and the retention of high-frequency details, we maintain the same design in the RGB branch whereas we turn off this option for depth and normal branches since high-frequency details are not desired for these two modalities. 
StyleGAN3 provides two configurations: StyleGAN3-T and StyleGAN3-R. Compared to StyleGAN3-T, StyleGAN3-R replaces $3 \times 3$ convolutions with $1\times 1$ convolutions and doubles the channel dimension to compensate for the lost capacity. Besides, it uses a radially symmetric $\mathrm{jinc}$-based downsampling filter to replace the $\mathrm{sin}$-based one. Though in practice, we noticed that StyleGAN3-R introduces many kaleidoscope-like patterns in results for the use of symmetric filters, this configuration achieves better performance on Stanford2D3D than StyleGAN3-T in terms of FID. Therefore, we use StyleGAN3-R by default. We follow StyleGAN3~\cite{karras2021styleganv3} to disable style mixing and path length regularization as they introduce extra difficulties in convergence for complex datasets. We also blur all modalities to discriminators using a Gaussian filter with $\sigma=10$ pixels over the first 200k images. As noted in~\cite{karras2021styleganv3}, this prevents the discriminator from focusing too heavily on high frequencies as training starts. 

For data augmentation operations, we follow ADA~\cite{ADA} to use pixel blitting ($x$-flip, $90^{\circ}$ rotations, integer translation), general geometric translations (isotropic scaling, arbitrary rotation, anisotropic scaling, fractional translation), color transformations (brightness, contrast, luma flip, hue rotation, saturation), image-space filtering and image-space corruptions (additive noise, cutout). Since only color transformations require different processing for different channels while other operations can work on individual channels, we only apply color transformations for the RGB modality.

\section{Implementation Details}
The original Stanford2D3D dataset~\cite{armeni2017stanford2d3d} provides training and validation splits. Training an unconditional generator does not explicitly require the split. Therefore, we use all data tuples from the dataset to train our Full models. The image resolution is up to $1024\times1024$. For quicker experiments, we resize all modalities to $256\times 256$. 
For RGB, it would be transformed to the range of [-1, 1]. The Stanford2D3D dataset stores depth within the range between 0 and 65,535. To cope with such large depth range, we perform max rescaling by:
\begin{equation}
    \mathrm{d' = \left(\frac{d - d.min()}{d.max()} - 0.5\right) * 2},
\end{equation} so that the ground-truth for the depth modality lies within [-1, 1]. The original data of depth contains large areas of blank holes, so we use the hole filling method~\cite{nyud} to inpaint the holes for better generation quality. The original Stanford2D3D dataset saves surface normal as RGB images and we maintain the same procedure as RGB images when dealing with normals. So the channels for RGB, depth and surface normal are 3, 1, and 3, respectively. 

We use a batch size of 4 for each GPU and 8 A100 GPUs using PyTorch 1.10.0 of CUDA 11.3 for most experiments. We use Adam optimizer~\cite{adam} to optimize both generator and discriminators with $\beta1 = 0$, $\beta2=0.99$ and $\epsilon=10^{-8}$. Following StyleGAN3~\cite{karras2021styleganv3}, we also use equalized learning rate for all trainable parameters~\cite{progressivegan}, minibatch standard deviation layer at the end of the discriminator~\cite{progressivegan}, exponential moving average of generator weights~\cite{progressivegan}, mixed-precision FP16/FP32 training~\cite{ADA} to facilitate training, R1 regularization~\cite{r1regularization}, and lazy regularization~\cite{karras2020styleganv2} to stabilize training. 

\section{More details about depth estimation and surface normal estimation}
We build {\rgbd} and {\rgbd*} comparison methods, use LeReS~\cite{yin2021depthestimation} to evaluate SIE and perform depth estimation using the combination of real and generated data. When building {\rgbd}, we pretrain LeReS~\cite{yin2021depthestimation} on the training set created in Sec.~5.3 while using the validation set to choose the best performing model as the final supervision model. While evaluating the SIE, we use its pretrained ResNeXt101 model\footnote{\url{https://cloudstor.aarnet.edu.au/plus/s/lTIJF4vrvHCAI31}}. We follow the provided training script\footnote{\url{https://github.com/aim-uofa/AdelaiDepth/tree/main/LeReS}} to train the model using the same configurations on 8 A100 GPUs for 30K iterations. We use the same procedure to train depth estimation models using real and generated data unless different number of training data. For {\rgbd*}, we use the same pretrained ResNeXt101 model as in evaluating SIE. 
When evaluating the metrics of surface normal estimation, we use pre-trained models downloaded using the official script~\cite{eftekhar2021omnidata,kar20223d}\footnote{\url{https://github.com/EPFL-VILAB/omnidata/blob/2a661c93285018b71141759d2a1ad53d8aed0e62/omnidata_tools/torch/tools/download_surface_normal_models.sh}}.

\section{More results}

\subsection{More Qualitative Results}
\label{sec:video_of_our_approach}

\begin{figure*}
\begin{center}
\includegraphics[width=0.85\linewidth]{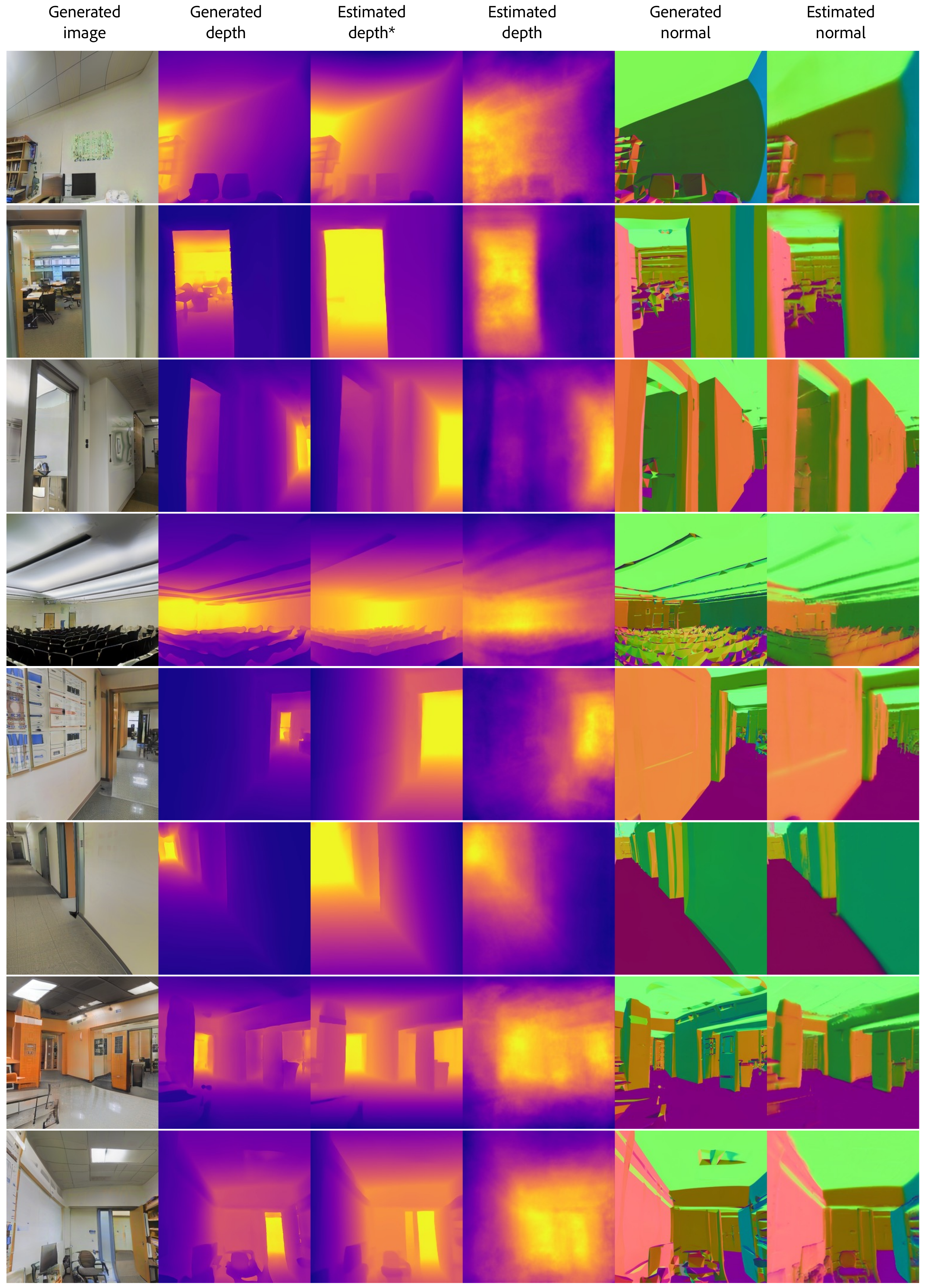}
\end{center}
\vspace{-6mm}
\caption{Demonstration of the comparison of our generated depth and surface normal with pre-trained depth estimators and normal estimator. The first column, second column and 5th column are generated RGB images, depths and normals from our model. The third column (Estimated depth*) represents the results estimated by pre-trained LeReS~\cite{yin2021depthestimation} by feeding the generated RGB images. The fourth column (Estimated depth) is the result obtained from LeReS which is trained on the in-domain dataset, namely Stanford2D3D. The final column (Estimated normal) is obtained by a pre-trained surface normal estimator~\cite{eftekhar2021omnidata,kar20223d} tested on the generated RGB images.}
\label{fig:benefit_of_multitask_learning}
\end{figure*}

In Sec.~4.3 of the manuscript, we show that our approach of training multiple modalities together gives better performance than {\rgbd} and {\rgbd*} which are trained on the RGB modality first and then move to depth using direct supervised learning. Aside from the difficulty of retrieving information from the features learned from RGB images for depth synthesis, an obvious drawback of this approach is the dependence on pre-trained models to provide supervisions. If the pre-trained models behave poorly on the target domain, learning from those models is not reliable. We show in Fig.~\ref{fig:benefit_of_multitask_learning} that when compared to pre-trained models, our generated depth (2nd column) and surface normal prediction (5th column) are better in capturing correct depth ranges and presenting clear layouts. The pre-trained depth estimator (3rd column) gives incorrect depth ranges and the surface normal estimator (final column) suffers from local blurriness. Even after we train the depth estimator on the whole Stanford2D3D dataset and then apply the depth estimator to provide supervision, the results (4th column) are still inferior to ours. In a sense, it is reasonable, since our approach directly learns from real ground-truth annotations while using pre-trained models (e.g., depth estimator) assumes learning from synthetic predicted annotations which could be unreliable.

We have two accompanying videos entitled ``\textbf{RGB+depth.mp4}'' and ``\textbf{RGB+depth+normal.mp4}'' under the ``\textbf{videos}'' folder to show the generated results of our model trained on \{RGB, depth\} and \{RGB, depth, normal\}, respectively. The results show the smoothness of transitions across different scenes and the consistency of different modalities. Even without clear guidance of camera poses, our model is still able to interpolate reasonably in a particular scene.

For each video clip, we sample 11 distinct latent codes, $z$, projecting them to corresponding intermediate latent codes, $w$. We perform a linear interpolation between two successive $w$s at a frame rate of 60, generating the associated RGB images and other modalities with the generator. The video is composed by compiling these sequential frames.


\subsection{Consistent RGBD Generation Video}


We have provided a video titled ``\textbf{consistent-rgbd-generation.mp4}'' in the ``\textbf{videos}'' directory to demonstrate the consistency between our generated depths and RGB images. Conversely, the depth estimations based on RGB images using the off-the-shelf depth estimator~\cite{yin2021depthestimation} present inconsistencies, either locally or globally, across frames.

\subsection{Cross-domain Fine-tuning Video}

As described in Sec.~5.2 of the manuscript, we can hold out a particular scene from the Stanford2D3D dataset, i.e., \emph{auditorium}, to pre-train our model and then fine-tune our model on the held-out scene by using only a few pairs and the rest RGB images. During fine-tuning, we observe that data augmentation for discriminators would cause incorrect layouts in the results, e.g., chairs in the same auditorium scene facing towards opposite directions. We assume when the number of pairs drastically decreases, it also decreases the chances for the discriminator to see real data and tuples. Hence, the distorted data is leaked to the generator, causing incorrect geometries. We hence remove the ADA for fine-tuning. These models are fine-tuned using around 10K training iterations, which is around 1/78 of pre-training iterations. 

We provide one accompanying video entitled ``\textbf{finetune.mp4}'' under the ``\textbf{videos}''
folder, which shows the result of the pre-trained model and then the models trained on different portions of pairwise data. Note that we use the same latent code $z$ to generate those videos and observe that though these are different models, the results at the same time step interestingly have related structures, indicating that after fine-tuning, some initial properties still remain. The transition is quite smooth and the quality is decent for the fine-tuning results, considering that we only use a small number of pairwise data.



\subsection{Using Only Generated Data for Depth Estimation}
\begin{figure}
\begin{center}
\includegraphics[width=\linewidth]{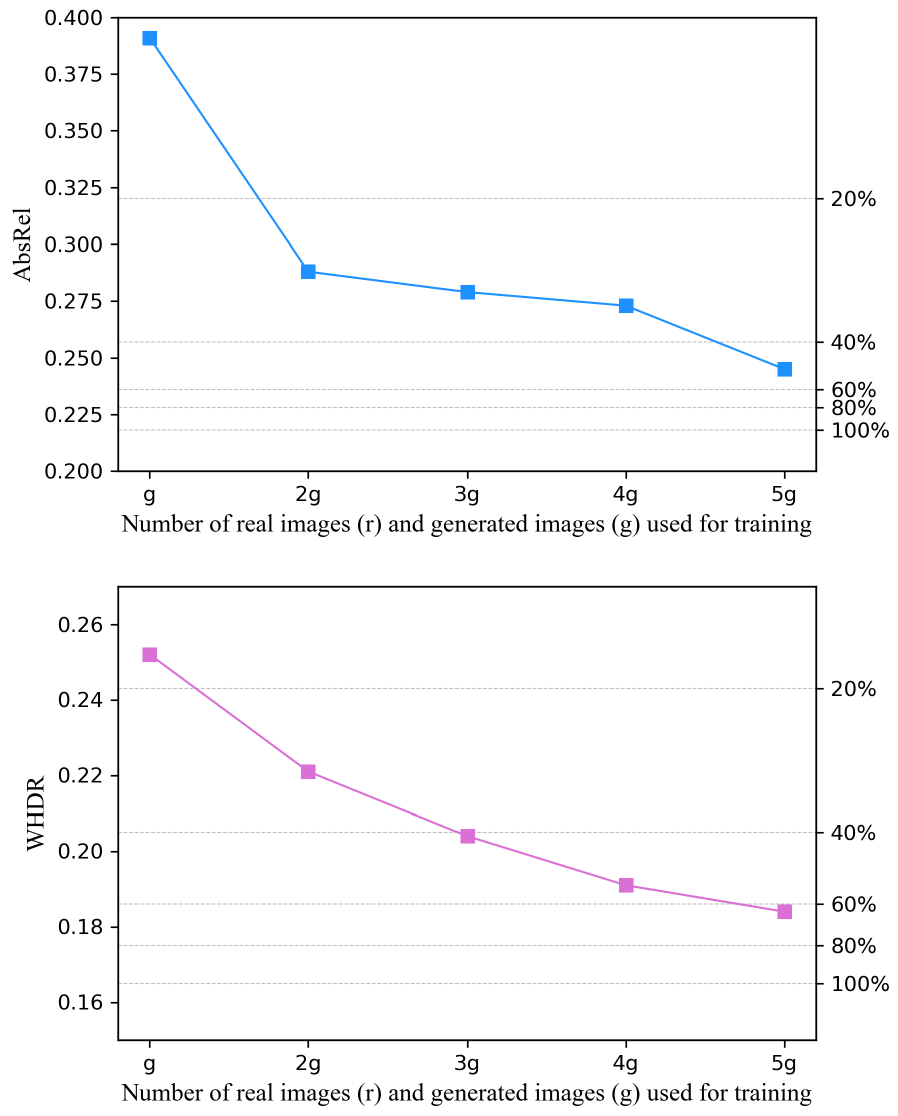}
\end{center}
\caption{Depth estimation performance of LeReS~\cite{yin2021depthestimation} when trained on different number of generated data. The $x$-axis represents different portions of generated images. The left $y$-axes are the numbers of AbsRel or WHDR obtained by the corresponding models, respectively. The dashed horizontal lines indicate the performances of training on different portions of training dataset. ``$g$'' represent the numbers of generated images. In this case, $g=12,045$, which is 20\% size of the training set.}
\label{fig:depth_estimation_pure_generated}
\end{figure}

In addition to the results of using both generated and real data (20\%) to train the depth estimation network~\cite{yin2021depthestimation}, we also experiment on using generated data only. As shown in Fig.~\ref{fig:depth_estimation_pure_generated}, using more generated data can get a steady improvement over both metrics. When the size of generated data equals to that of real data ($5g$=100\% real training data), the performance is close to using 60\% of real training data. This promising result indicates the possibility of using our model to generate datasets for learning downstream tasks.

\end{document}